\documentclass[preprint]{sig-alternate-05-2015}

\usepackage[utf8]{inputenc} 
\usepackage[T1]{fontenc}    
\usepackage[numbers,sort&compress]{natbib}
\usepackage{blindtext, graphicx}
\usepackage{tikz}
\usepackage{pgfplots}
\pgfplotsset{compat=1.3}
\usetikzlibrary{intersections}
\usepackage{hyperref}       
\usepackage{url}            
\usepackage{booktabs}       
\usepackage{tabularx}
\usepackage{amsfonts}       
\usepackage{nicefrac}       
\usepackage{microtype}      
\usepackage{subcaption}
\usepackage{todonotes}
\usepackage{glossaries}
\usepackage{subcaption}
\usepackage[font=small,skip=1pt]{caption}
\usepackage{soul}
\usepackage{longtable}
\usepackage{multirow}
\usepackage{rotating}

\newcommand{\dblquotes}[1]{``#1''}
\newcommand{\OurScheme}{\textsc{Finn}}
\newcommand*\rfrac[2]{{}^{#1}\!/_{#2}}
\newcommand{\mlp}{\ensuremath{\mathrm{mlp}}}
\newcommand{\mlps}[1]{\ensuremath{\mlp{}(#1)}}
\newcommand{\mlpf}[2]{\ensuremath{\mlps{\rfrac{#1}{#2}}}}
\newcommand{\cnn}{\ensuremath{\mathrm{cnn}}}
\newcommand{\cnns}[1]{\ensuremath{\cnn{}(#1)}}
\newcommand{\cnnf}[2]{\ensuremath{\cnns{\rfrac{#1}{#2}}}}
\newcommand{\cnnnp}{\ensuremath{\cnn_{\mathrm{NoPad}}}}
\newcommand{\cnnnps}[1]{\ensuremath{\cnnnp{}(#1)}}
\newcommand{\cnnnpf}[2]{\ensuremath{\cnnnps{\rfrac{#1}{#2}}}}
\newlength{\subfigpad}
\newlength{\figpad}
\newif\ifkultra
\kultratrue
\ifkultra
\newcommand{\LargeFpgaName}{KU115}

\newcommand{\LargeFpgaFloatPeak}{483~\gls{GFLOPS}}
\newcommand{\LargeFpgaBinPeak}{46~\gls{TOPS}}
\newcommand{\LargeFpgaPlatform}{ADM-PCIE-8K5}

\else 
\newcommand{\LargeFpgaName}{XC7VX690T}

\newcommand{\LargeFpgaFloatPeak}{252~\gls{GFLOPS}}
\newcommand{\LargeFpgaBinPeak}{24~\gls{TOPS}}
\newcommand{\LargeFpgaPlatform}{ADM-PCIE-7V3}

\fi

\tikzset{
	declare function={
		mysign(\x) = (and(\x<=0, 1) * -1) +
		       (and(\x>0, 1) * 1);
		}
	}

\glsdisablehyper
\newacronym{FPGA}{FPGA}{field programmable gate array}
\newacronym{CNN}{CNN}{convolutional neural network}
\newacronym{BNN}{BNN}{binarized neural network}
\newacronym{MLP}{MLP}{multilayer perceptron}
\newacronym{OFM}{OFM}{output feature map}
\newacronym{IFM}{IFM}{input feature map}
\newacronym{OCM}{OCM}{on-chip memory}
\newacronym{MVU}{MVTU}{Matrix--Vector--Threshold Unit}
\newacronym{SWU}{SWU}{Sliding Window Unit}
\newacronym{TU}{TU}{Thresholding Unit}
\newacronym{PU}{PU}{Pooling Unit}
\newacronym{II}{II}{initiation interval}
\newacronym{PE}{PE}{Processing Element}
\newacronym{NN}{NN}{neural network}
\newglossaryentry{FPS}{name=FPS,plural=FPS,description=frames per second}
\newacronym{HLS}{HLS}{High-Level Synthesis}
\newacronym{ILSVRC}{ILSVRC}{ImageNet Large Scale Visual Recognition Competition}
\newacronym[plural=TOPS,longplural=trillion operations per second]{TOPS}{TOPS}{trillion operations per second}
\newacronym[plural=GOPS,longplural=billion operations per second]{GOPS}{GOPS}{billion operations per second}
\newacronym[plural=GOP,longplural=billion operations]{GOP}{GOP}{billion operations}
\newacronym[plural=GFLOP,longplural=billion floating point operations]{GFLOP}{GFLOP}{billion floating point operations}
\newacronym[plural=GFLOPS,longplural=billion floating point operations per second]{GFLOPS}{GFLOPS}{billion floating point operations per second}
\newacronym{MAC}{MAC}{multiply accumulate}
\newglossaryentry{LUT}{name=LUT,description=lookup table}

\newcommand{\PowerWall}{$P_{\mathrm{wall}}$}
\newcommand{\PowerFPGA}{$P_{\mathrm{chip}}$}

\newcommand{\FINN}{\OurScheme{}}
\newcommand{\approxtilde}[1]{{\raise.17ex\hbox{$\scriptstyle\sim$}}#1}

\title{Scaling Binarized Neural Networks on Reconfigurable Logic}

\numberofauthors{1} 
\author{
Nicholas J. Fraser\textsuperscript{*\ddag}, Yaman Umuroglu\textsuperscript{*\dag}, Giulio Gambardella\textsuperscript{*}, Michaela Blott\textsuperscript{*},\\
Philip Leong\textsuperscript{\ddag}, Magnus Jahre\textsuperscript{\dag} and Kees Vissers\textsuperscript{*}\\
       \affaddr{\textsuperscript{*}Xilinx Research Labs};
       \affaddr{\textsuperscript{\dag}Norwegian University of Science and Technology};
       \affaddr{\textsuperscript{\ddag}University of Sydney}\\
       \email{nfraser@xilinx.com, yamanu@idi.ntnu.no}
}

\toappear{To appear in the PARMA-DITAM workshop at HiPEAC 2017, January 2017.}

\begin{document}


\maketitle

\begin{abstract}
\Glspl{BNN} are gaining interest in the deep learning community due to their significantly lower computational and memory cost.
They are particularly well suited to reconfigurable logic devices, which contain an abundance of fine-grained compute resources and can result in smaller,
lower power implementations, or conversely in higher classification rates.
Towards this end, the \OurScheme{} framework was recently proposed for building fast and flexible \gls{FPGA} accelerators for \glspl{BNN}.
\OurScheme{} utilized a novel set of optimizations that enable efficient mapping of \glspl{BNN} to hardware and implemented fully connected,
non-padded convolutional and pooling layers, with per-layer compute resources being tailored to user-provided throughput requirements.
However, FINN was not evaluated on larger topologies due to the size of the chosen \gls{FPGA}, and exhibited decreased accuracy due to lack of padding.
In this paper, we improve upon \OurScheme{} to show how padding can be employed on \glspl{BNN} while still maintaining a 1-bit datapath and high accuracy.
Based on this technique, we demonstrate numerous experiments to illustrate flexibility and scalability of the approach.
In particular, we show that a large BNN requiring 1.2 billion operations per frame running on an \LargeFpgaPlatform{} platform can classify images at 12~k\gls{FPS} with 671~$\mu$s latency while drawing less than 41 W board power and classifying CIFAR-10 images at 88.7\% accuracy.
Our implementation of this network achieves 14.8 trillion operations per second.
We believe this is the fastest classification rate reported to date on this benchmark at this level of accuracy.
\end{abstract}

\section{Introduction}
\label{sec:introduction}
\Glspl{CNN} provide impressive classification accuracy in a number of application domains,
but at the expense of large compute and memory requirements~\cite{mnist}.
A significant body of research is investigating compression techniques combining numerous approaches such as:
weight and synapse pruning;
data compression techniques such as quantization, weight sharing and Huffman coding; and
reduced precision with fixed point arithmetic~\cite{han2015deep, iandola2016squeezenet, iandola2015firecaffe}. 
Recently, an extreme form of reduced precision networks, known as \glspl{BNN}~\cite{binarynet}, have gained significant interest as they can be implemented for inference at a much reduced hardware cost.
This is due to the fact that multipliers and accumulators become XNORs and popcounts respectively, and both are significantly lighter in regards to resource and power footprint.
For example, a \LargeFpgaName{} offers \LargeFpgaFloatPeak{} compared to \LargeFpgaBinPeak{} for binary synaptic operations.
This is visualized in the roofline models in Figure~\ref{fig:roofline} which illustrates theoretical peak performance for numerous reduced precision compute operations.%
\footnote{Assuming 70\% device utilization, 250 MHz clock frequency and 178 LUTs and 2 DSPs per average floating point operation, and 2.5 LUTs per binary XNOR-popcount operation.}
Furthermore, the model size is greatly reduced and typically small enough to fit in \gls{OCM}, again reducing power, simplifying the implementation and providing much greater bandwidth.

\OurScheme{}~\cite{finn} describes a framework for mapping \glspl{BNN} to reconfigurable logic.
However, it focuses on \glspl{BNN} for embedded applications and as such, the results reported are for smaller network sizes running on an embedded platform.
In this work, we briefly summarise \OurScheme{} and analyse it from the perspective of scaling to larger networks and devices, such as those targeted for data centers.
Firstly, we focus on several technical issues that arise when scaling networks on \OurScheme{} including: BRAM usage, throughput limitations and resource overheads.
We also identify several properties of \gls{CNN} layers which make them map to \OurScheme{} more efficiently.
Our results, measured on an \LargeFpgaPlatform{} platform~\cite{ku115}, show that indeed very high image classification rates, minimal latency with very high power efficiency can be achieved by mapping \glspl{BNN} to \glspl{FPGA}, even though improvements may be made.
Secondly, we highlight an issue of padding, a common feature of large \glspl{CNN}, which may cause significant hardware overheads.
We propose an alternative form of padding, which maps more efficiently to reconfigurable logic.
Specifically, the contributions of this work are:
1) measured performance results for large-scale networks on an \LargeFpgaPlatform{} board;
2) an analysis of \OurScheme{} for large-scale problems, highlighting some bottlenecks as well as proposing solutions; and
3) a form of padding, which achieves high accuracy while also maintaining a binary datapath.

\section{Background}
\label{sec:background}
A great deal of prior work on mapping neural networks to hardware exist for \glspl{FPGA}, GPUs and ASICs to help increase inference rate or improve energy efficiency.
We refer the reader to the work by Misra and Saha~\cite{surveyannhw} for a comprehensive survey of prior works.
In general we distinguish four basic architectures:
1) a \textit{single processing engine}, usually in the form of a \textit{systolic array}, which processes each layer sequentially~\cite{ovtcharov2015accelerating,zhang2015optimizing,chen2016eyeriss,YodaNN};
2) a \textit{streaming architecture}~\cite{venieris2016fpgaconvnet,alemdar2016ternary}, consisting of one processing engine per network layer;
3) a \textit{vector processor}~\cite{farabet2009cnp} with instructions specific to accelerating the primitives operations of convolutions; and
4) a \textit{neurosynaptic processor}~\cite{esser2016convolutional}, which implements many digital neurons and their interconnecting weights.
Significant research investigates binarization of neural networks whereby either input activations, synapse weights or output activations or a combination thereof are binarized.
If all three components are binary, we refer to this as \textit{full binarization}~\cite{bitwiseneuralnet}.
If not all three components are binary, we refer to this as \textit{partial binarization}.
The seminal XNOR-Net work by Rastegari~et~al.~\cite{xnornet} applies convolutional \glspl{BNN} on the ImageNet dataset with topologies inspired by AlexNet, ResNet and GoogLeNet, reporting top-1 accuracies of up to 51.2\% for full binarization and 65.5\% for partial binarization.
DoReFa-Net by Zhou~et~al.~\cite{dorefa} explores reduced precision with partial and full binarization on the SVHN and ImageNet datasets, including best-case ImageNet top-1 accuracies of 43\% for full and 53\%  for partial binarization.
Finally, the work by Courbariaux~et~al.~\cite{binarynet} describes how to train fully-connected and convolutional networks with full binarization and batch normalization layers, reporting competitive accuracy on the MNIST, SVHN and CIFAR-10 datasets.
All \glspl{BNN} used in this work are trained by a methodology based on the one described by Courbariaux~et~al.~\cite{binarynet}, and unset bits represent a numerical -1 value while set bits represent a +1.
The downside to the high performance characteristics of \glspl{BNN} is a small drop in accuracy, in comparison to floating point networks.
Improving the accuracy for reduced precision \glspl{CNN} is an active research area in the machine learning community and first evidence shows that accuracy can be improved by increasing network sizes~\cite{ResiliencyUnderQuantization}.

\section{\glspl{BNN} on Reconfigurable Logic}
\label{sec:architecture}

\begin{figure}
	\centering
    \hspace{\subfigpad}
	\begin{subfigure}[b]{\linewidth}
		\centering
		\includegraphics[height=3cm]{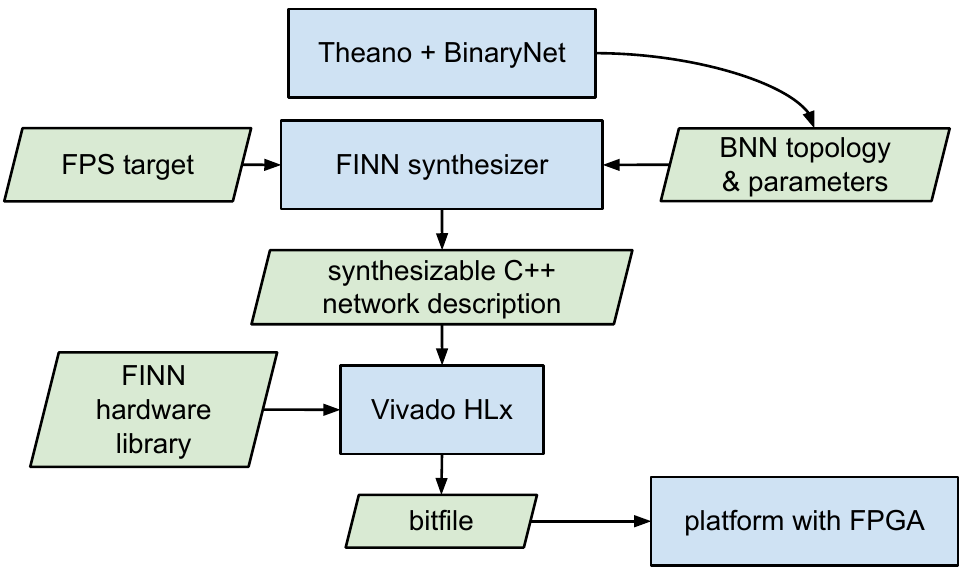}
		\caption{Accelerator generation.}
		\label{fig:workflow-summary}
	\end{subfigure}

    \hspace{\subfigpad}
	\begin{subfigure}[b]{\linewidth}
		\centering
		\includegraphics[height=2.7cm]{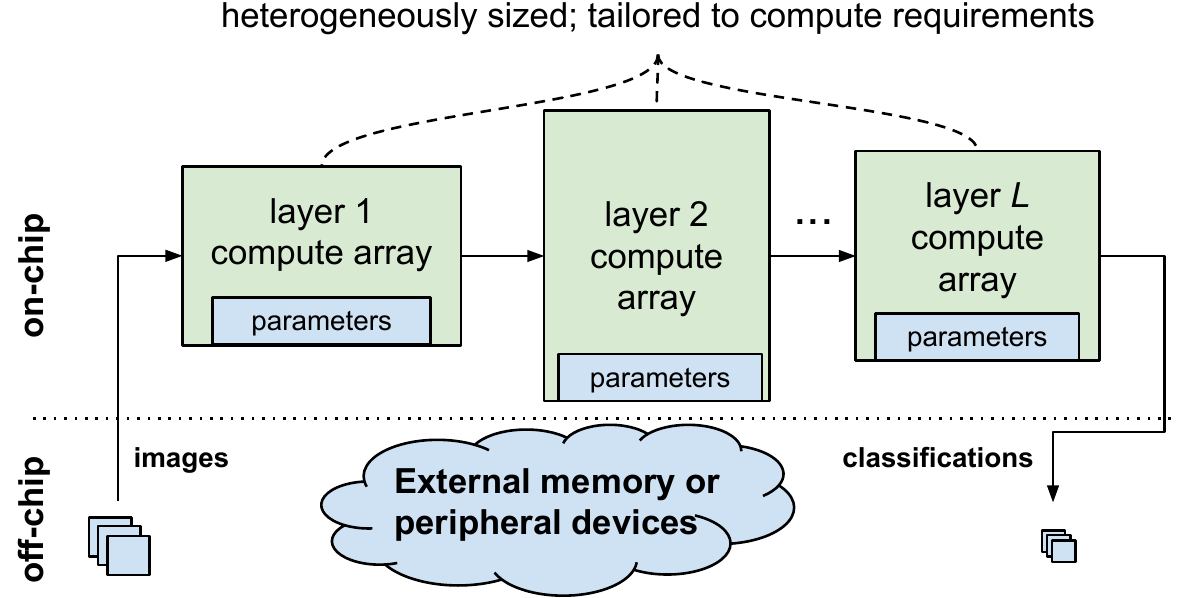}
		\caption{Top-level architecture.}
		\label{fig:arch}
	\end{subfigure}
    \vspace{0.5ex}

    \hspace{\subfigpad}
	\begin{subfigure}[b]{0.48\linewidth}
		\centering
		\includegraphics[height=2cm]{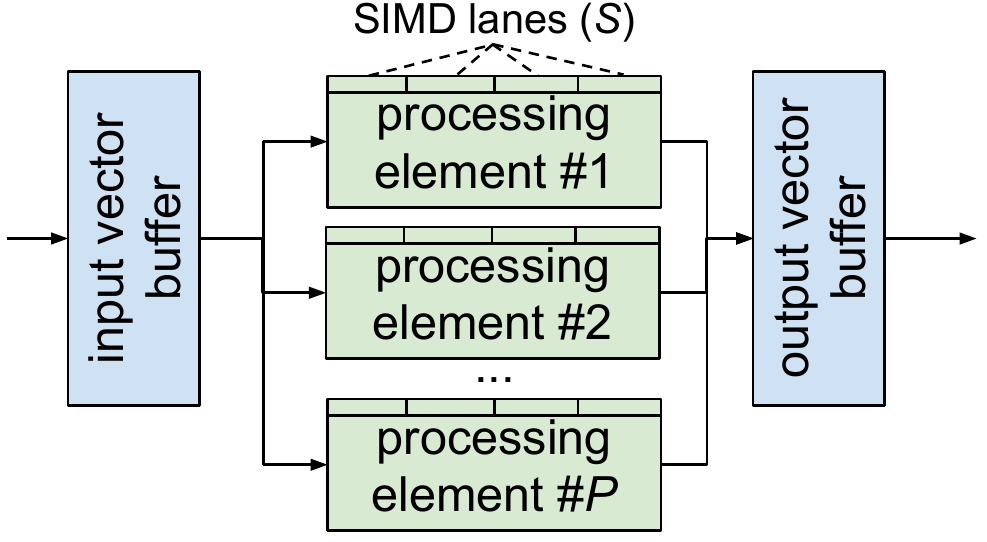}
		\caption{Building block (MVTU).}
		\label{fig:mvtu}
	\end{subfigure}
    \hspace{\subfigpad}
	\begin{subfigure}[b]{0.48\linewidth}
		\centering
		\includegraphics[height=2cm]{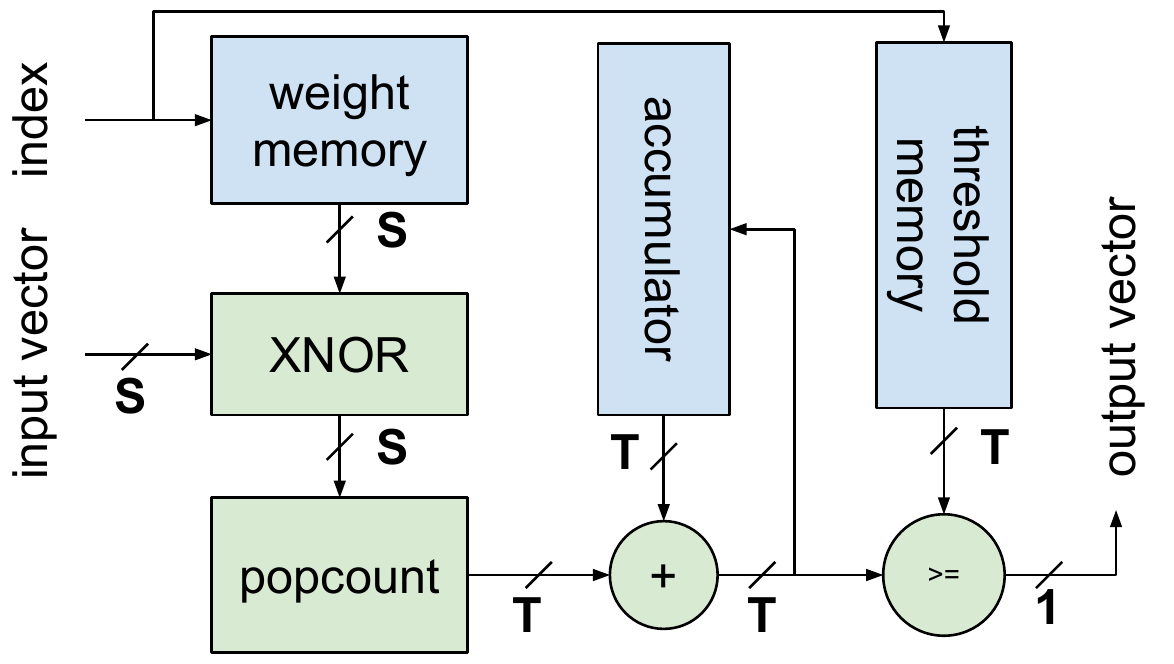}
		\captionof{figure}{MVTU datapath.}
		\label{fig:pe-datapath}
	\end{subfigure}
    \vspace{0.5ex}
	\caption{\FINN{} workflow and architecture, reproduced from \cite{finn}.}
	\label{fig:finn}
    \vspace{\figpad}
\end{figure}

This work builds on top of \FINN{} \cite{finn}, a framework for building scalable and fast \gls{BNN} inference accelerators on \glspl{FPGA}.
\FINN{} is motivated by observations on how \glspl{FPGA} can achieve performance in the \gls{TOPS} range using XNOR--popcount--threshold datapaths to implement the \glspl{BNN} described by Courbariaux et al.~\cite{binarynet}.
Given a trained \gls{BNN} and target frame rates, \FINN{} follows the workflow in Figure \ref{fig:workflow-summary} to compose a \gls{BNN} accelerator from hardware building blocks.
In more detail, a given network topology and model retrieved through Theano~\cite{theano}, together with design targets in form of resource availability and classifcation rate, is processed by the synthesizer which determines the scaling settings and produces a synthesizable C++ description of a heterogeneous streaming architecture.%
\footnote{To achieve portability, we chose a commercial high level synthesis tool, Vivado \gls{HLS}~\cite{vivadohls}, for the implementation. The tool enables faster development cycles via high-level abstractions, and provides automated pipelining to meet the clock frequency target.}
The top-level architecture is exemplified in Figure \ref{fig:arch} and has two key differentiators compared to prior work on \gls{FPGA} \gls{CNN} accelerators.
First, all \gls{BNN} parameters are kept in \gls{OCM}, which greatly increases arithmetic intensity, reduces power and simplifies the design. 
Furthermore, one streaming compute engine is instantiated per layer, with resources tailored to fit each layer's compute requirements and the user-defined frame rate.
Compute engines communicate via on-chip data streams and each produces and consumes data in the same order with the aim of minimizing buffer requirements in between layers.
Thereby each engine starts to compute as soon as the previous engine starts to produce output.
In essence, we build a custom architecture for a given topology rather than scheduling operations on top of a fixed architecture, as would be the case for typical systolic array based architectures, and avoid the \dblquotes{one-size-fits-all} inefficiencies and reap more of the benefits of reconfigurable computing.

\subsection{The Matrix--Vector--Threshold Unit}
In more detail, the key processing engine in \FINN{} is the \gls{MVU} as illustrated in Figure \ref{fig:mvtu}, which computes binarized matrix-vector products and compares against a threshold to generate a binarized activation.
Convolutions are \emph{lowered}~\cite{chellapilla2006high} to matrix--matrix multiplications, using \gls{SWU} (described further in Section \ref{sec:PaddingImpl}) to generate the image matrix and the \gls{MVU} to carry out the actual arithmetic.
The \gls{SWU} generates the same vectors as those in~\cite{chellapilla2006high} but with the elements of the vector interleaved to reduce and simplify memory accesses and to avoid the need for data transposition between layers.
Internally, the \gls{MVU} consists of an input and output buffer, and an array of $P$ \glspl{PE}, shown in Figure~\ref{fig:pe-datapath}, each with a number of SIMD lanes, $S$.
The synapse weight matrix to be used is kept in \gls{OCM} distributed between \glspl{PE}, and the input images stream through the \gls{MVU} as each one is multiplied with the matrix.
Each \gls{PE} receives exactly the same control signals and input vector data, but multiply-accumulates the input with a different part of the matrix.
A \gls{PE} can be thought of as a hardware neuron capable of processing $S$ synapses per clock cycle.
Finally, the \gls{MVU} architectural template can also support partial binarization for non-binarized outputs and inputs.
Removing the thresholding stage provides non-binarized outputs, while using regular multiply-add instead of XNOR-popcount can handle non-binarized inputs.
These features are used in the first and last layers of networks that process non-binary input images or do not output a one-hot classification vector.

\subsection{Folding}
Depending on the use case, a neural network inference accelerator may have different throughput requirements in terms of the images classified per second (FPS).
In FINN, FPS is controlled by the per-layer parameters $P$ (number of PEs in an MVTU) and $S$ (number of SIMD lanes in each PE).
If the number of synapses, $Y$, connected to a neuron is greater than $S$, then the computation is \textit{folded} across the \gls{PE}, with the resulting \gls{PE} producing an activation every $F^s = Y / S$ clock cycles.
Similarly, if the number of neurons, $X$, in a layer exceeds $P$, then each \gls{PE} is responsible for calculating activations for $F^n = X / P$ neurons.
In total, it would take the \gls{MVU} $F^s \cdot F^n$ clock cycles to compute all its neuron activations.
The \glspl{MVU} are then rate balanced by adjusting their $P$ and $S$ values to match the number of clock cycles it takes to calculate all required activations for each layer.
As this is a balanced streaming system, the classification throughput $\mathrm{FPS}$ will be approximately $F_{\mathrm{clk}} / \mathit{II}$, where $F_{\mathrm{clk}}$ is the clock frequency, and the $\mathit{II}$ (Initiation Interval) is equal to the total folding factor $F^{\mathrm{tot}}=F^s \cdot F^n$ cycles for a fully-connected layer.
Note that convolutional layers have an extra folding factor, $F^m$, which is the number of matrix--vector products which need to be computed,
i.e., the number of pixels in a single \gls{OFM}.
Therefore, for convolutional layers the total folding factor is: $F^{\mathrm{tot}}=F^s \cdot F^n \cdot F^m$.


\subsection{BNN-specific Operator Optimizations}
\label{sec:Optimizations}
The methodology described in \cite{binarynet} forms the basis for training all \glspl{BNN} in this paper.
Firstly, in regards to arithmetic, we are using 1-bit values for all input activations, weights and output activations (full binarization), where an unset bit represents -1 and a set bit represents +1.
Binary dot products result in XNORs with popcounts (which count the number of set bits instead of accumulation with signed arithmetic). 
Secondly, all \gls{BNN} layers use batch normalization \cite{batchnorm} on convolutional or fully connected layer outputs, then apply the sign function to determine the output activation.
In \cite{finn} it is shown how the same output can be computed via thresholding, which combines the bias term, batch normalization and activation into a single function.
Finally, the networks described in \cite{binarynet} perform pooling prior to activations, i.e. pooling is performed on non-binarized numbers, which are then batch normalized and fed into the activation function.
However, as shown in~\cite{finn}, pooling can be equally performed after activation, once binarized, in which case it can be effectively implemented with the Boolean OR-operator.

\section{Padding for BNN Convolutions}
\label{sec:padding}
This section describes the improvements made to \OurScheme{} in this work.

\subsection{Padding using nonzero values}
\begin{figure}
	\centering
	\includegraphics[height=3.5cm]{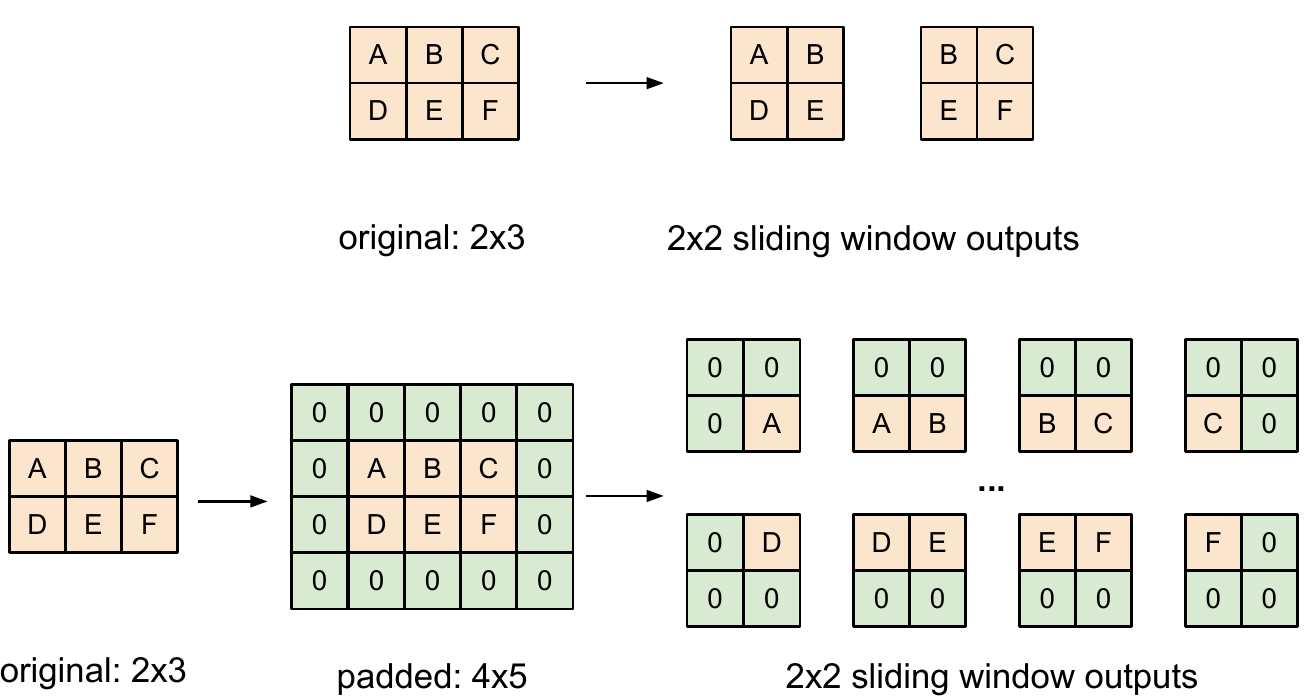}
	\caption{Convolution without (top) and with (bottom) padding.}
	\label{fig:PaddingExample}
    \vspace{\figpad}
    \vspace{2ex}
\end{figure}

Zero-padding is commonly applied for convolutional layers in deep neural networks, in order to prevent the pixel information on the image borders from being "washed away" too quickly \cite{cs231n}.
Figure \ref{fig:PaddingExample} illustrates the sliding window outputs on the same image with and without padding.
Observe that the pixels on the border (such as A and F) occur more frequently in the sliding window outputs when padding is used, thus preventing them from being "washed away" too quickly in the next layer.

A challenge arises for zero-padding in the context of \glspl{BNN} with only $\{-1, +1\}$ arithmetic: there is no zero value defined.
In fact, the original BinaryNet~\cite{binarynet} paper uses ternary values $\{-1, 0, +1\}$ for the forward pass, with zeros used for padding.
However, ternary values require two bits of storage, essentially doubling the \gls{OCM} required to store values and the bitwidth of the datapath.
Since \OurScheme{} focuses on \glspl{BNN} that fit entirely into on-chip memory of a single \gls{FPGA}, minimizing the resource footprint is essential.
Thus, a padding solution that avoids ternary values is preferable.
A straightforward solution would be to use e.g. -1 as the padding value, and expect that the \gls{BNN} learns weights which compensate for these values.
Surprisingly, -1-padding works just as well as 0-padding according to our results, which are presented in Section \ref{sec:PaddingResults}.


\subsection{Streaming padding for FINN}
\label{sec:PaddingImpl}
\begin{figure}
	\centering
	\includegraphics[height=3.5cm]{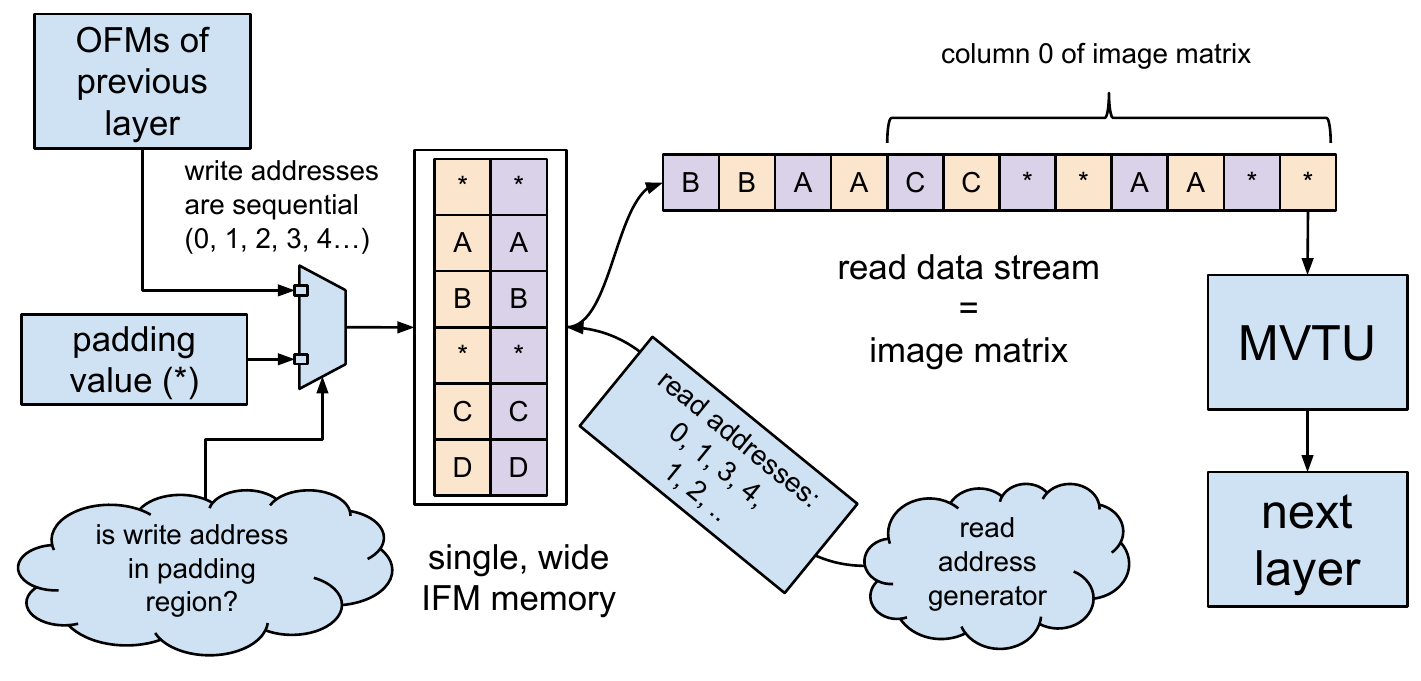}
	\caption{\OurScheme{} SWU enhanced with streaming padding.}
	\label{fig:PaddingImpl}
    \vspace{\figpad}
\end{figure}
\OurScheme{} lowers~\cite{chellapilla2006high} convolutions to matrix-matrix multiplication of the filter weight matrix with the image matrix.
The image matrix is generated on-the-fly by the \gls{SWU}.
Figure \ref{fig:PaddingImpl} illustrates how the \OurScheme{} \gls{SWU} is enhanced to support streaming padding for convolution layers.
The key operational principle is the same as in \OurScheme{}.
Namely, a single, wide \gls{IFM} memory is used to store the feature maps into \gls{OCM} in the order they arrive, and the addresses that correspond to the sliding window pixels are read out.
Padding is achieved by a multiplexer that chooses the data source for writing into the \gls{IFM} memory.
If the current write address falls into the padding region, the padding value (e.g. -1) is written into the memory; otherwise, an element from the output stream of the previous layer is written instead.

\section{Evaluation}
\label{sec:results}

\subsection{Experimental Setup}
\subsubsection{BNN Topologies}
The network topologies used for our experiments are all based on the \gls{CNN} topology described in \cite{binarynet}, which we denote as \cnn{}.
This topology is inspired by the VGG16 network~\cite{simonyan2014very}, which consists of three groups of (3x3 convolution -- 3x3 convolution -- 2x2 maxpooling) layers, and two fully-connected layers at the end.
To explore how \OurScheme{} performs on a range of network sizes,
we introduce a scaling factor, $\sigma$, to scale the width of each layer, and denote the resulting topology as \cnns{\sigma}.
Note that $\sigma$ does not influence the number of layers in a network, it merely affects:
1) the number of neurons in each fully connected layer; and
2) the number of filters in each convolutional layer.
Specifically, \cnns{0.5} has half as many filters in each convolutional layer and half as many neurons in each fully connected layer, compared to the \gls{CNN} described in \cite{binarynet}.
In terms of convolutional networks, ~\cite{finn} only evaluated a single non-padded \gls{BNN} topology (\cnnnpf{1}{2}).
In this work, we consider \cnnf{1}{2} as well as smaller (\cnnf{1}{4}) and bigger (\cnns{1}) padded convolutional topologies to investigate how \OurScheme{} scales.

In order to simulate a realistic use case, we consider an application with a fixed \gls{FPS} requirement, i.e., real-time object recognition of a video stream.
If one considers an 800 $\times$ 600 video stream at 25 \gls{FPS}, which partitioned into tiles of 32 $\times$ 32 for classification.
In order to classify the tiles in real-time, a classification rate of approximately 12 k\gls{FPS} would be required.
We use this image rate as our target for all experiments and adjust the number of \glspl{PE} and SIMD accordingly in each layer of each design.

\subsubsection{The Platform}
The target board is an Alpha Data \LargeFpgaPlatform{} which features a Xilinx Kintex UltraScale XCKU115-2-FLVA1517E FPGA (KU115).
The KU115 offers 663k LUTs, 2160 BRAMs (36k) and 5520 DSPs and is running at 125 MHz for our experiments.
The host machine is a IBM Power8 8247-21L with 80 cores at 3.69 GHz and 64 GB of RAM and it is running Ubuntu 15.04.
In all experiments, all parameters are stored in \gls{OCM} while the test images and the predicted labels are read from and written to the host memory directly.
The provided resource counts include the PCI Express infrastructure used for moving data streams as well as the BNN accelerator.
Although we are not able to provide per-experiment power measurements, the maximum power consumption observed for this board was 41~W on a board power dissipation benchmark test, and we expect that the real power dissipation values for BNN accelerators will be significantly lower than this.

\subsection{Effects of Padding}
\label{sec:PaddingResults}
\begin{table}
\centering
\caption{Accuracy with different padding modes for CIFAR-10.}
%
\scriptsize
\begin{tabular}{cc|ccc}
\toprule
& & \multicolumn{3}{c}{\emph{Padding Mode}} \\
& & no-padding & 0-padding & -1-padding \\ 
 \midrule
\multirow{3}{*}{\begin{sideways}\emph{Scale}\end{sideways}} &
$\sigma=\rfrac{1}{4}$   & 75.6\%    & 78.2\%    & 79.1\%      \\ 
& $\sigma=\rfrac{1}{2}$ & 80.1\%    & 85.2\%    & 85.2\%    \\ 
& $\sigma=1$            & 84.2\%    & 88.6\%    & 88.3\%    \\ 
\bottomrule
\end{tabular}

\label{tab:PaddingResults}
\end{table}

To investigate how different padding modes affect accuracy, we trained a set of convolutional \glspl{BNN} on the CIFAR-10 dataset with different scaling factors ($\sigma$).
The convolutions used are 3$\times$3, so one pixel of padding is added on each border.
The results are summarized in Table \ref{tab:PaddingResults}.
As expected, using 0-padding improves accuracy by 4-5\% compared to no-padding, indicating that the conventional wisdom on padding increasing accuracy also applies to \glspl{BNN}.
Furthermore, we can see that the accuracy of -1-padded networks are on par with the 0-padded ones of same scale.
This suggests that \glspl{BNN} are able to learn to compensate for the -1 values used for padding by adjusting the weight values and thresholds, and the accuracy benefits can be still obtained with a binary (as opposed to ternary) datapath.

\begin{table}
\centering
\caption{Operations per image with different padding modes for CIFAR-10.}
%
\scriptsize
\begin{tabular}{cc|ccc}
\toprule
& & \multicolumn{3}{c}{\emph{Padding Mode}} \\
& & no-padding & 0-padding & -1-padding \\ 
 \midrule
\multirow{3}{*}{\begin{sideways}\emph{Scale}\end{sideways}} &
$\sigma=\rfrac{1}{4}$ & 30.4~M & 78.5~M & 78.5~M \\ 
& $\sigma=\rfrac{1}{2}$ & 118.9~M & 310.3~M & 310.3~M \\ 
& $\sigma=1$ & 530.1~M & 1234.1~M & 1234.1~M \\ 
\bottomrule
\end{tabular}

\label{tab:PaddingOpCountResults}
\vspace{-3ex}
\end{table}

It should also be noted that no-padding results in a significant reduction in the amount of operations per frame and the number of parameters.
Thus, it is worthwhile to examine the computation versus accuracy tradeoffs in the context of padding.
Table \ref{tab:PaddingOpCountResults} lists the total number of XNOR-popcount operations necessary to classify one image using different padding modes and scaling factors.
We can observe that the no-padding topology variant for the same scale factor requires $2-3\times$ less computation.
However, this comes at a cost of higher error rate, and a smaller-but-padded network may be advantageous over a larger-but-not-padded network.
For instance, \cnnf{1}{4} classifies at 79\% accuracy using 78.5~M operations, whereas the \cnnnpf{1}{2} classifies at 80.1\% accuracy using 118.9~M operations.
Thus, \cnnf{1}{4} may be preferable due to its lower computational cost if a 1\% drop in accuracy is acceptable for the use case at hand.

\subsection{Scaling to Larger Networks}
\begin{figure}
\centering
\includegraphics{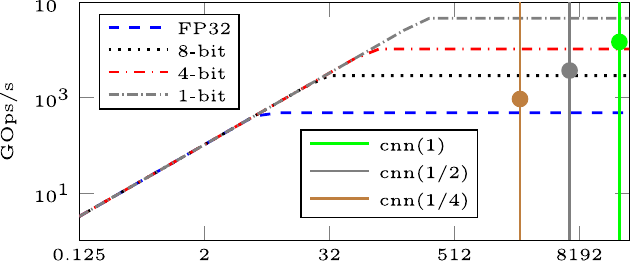}
\captionof{figure}{\LargeFpgaName{} roofline with different datatypes.}
\label{fig:roofline}
\vspace{\figpad}
\end{figure}

\begin{table}
	\centering
	\caption{Key performance and resource utilization results achieved by this work (top) and \OurScheme{} (bottom) on a number of \gls{BNN} topologies.}
	\newcommand{\tss}[1]{\textsuperscript{#1}} 
\newcommand{\pf}{\PowerFPGA{}}
\newcommand{\pw}{\PowerWall{}}
\resizebox{\linewidth}{!}{
\scriptsize
\begin{tabular}{llrrrrrr}
\toprule
&Network        & Device & LUT    & BRAM & kFPS   & GOps/s \\
\midrule
\ifkultra
\multirow{3}{*}{\begin{sideways}Padded\end{sideways}}&\cnnf{1}{4}    & KU115  & 35818  & 144  & 12.0   &  938\\
&\cnnf{1}{2}    & KU115  & 93755  & 386  & 12.0   & 3,711\\
&\cnns{1}       & KU115  & 392947 & 1814 & 12.0   & 14,814\\
\else
&\cnnf{1}{4}    & VX690T & 35818  & 144  & 12.0   & 938\\
&\cnnf{1}{2}    & VX690T & 93755  & 386  & 12.0   & 3,711\\
&\cnns{1}       & VX690T & 159735 & 1356 & 6.35   & 7,839\\
\fi
\midrule
\multirow{4}{*}{\begin{sideways}FINN~\cite{finn}\end{sideways}}&\cnnnpf{1}{2}  & Z7045  & 54538  & 192  & 21.9   & 2,466 \\
&\mlpf{1}{16}   & Z7045  & 86110  & 130.5& 12,361 &  8,265 \\
&\mlpf{1}{8}    & Z7045  & 104807 & 516.5& 6,238  & 11,613 \\
&\mlpf{1}{4}    & Z7045  & 79097  & 398  & 1,561  & 9,086 \\

\bottomrule
\ifkultra
\fi
\end{tabular}
}
\let\tss\undefined
\let\pf\undefined
\let\pw\undefined

	\label{tab:finn-key-results}
\end{table}
A results summary is shown in Table~\ref{tab:finn-key-results} which also shows the accuracy achieved by the implemented networks on a number of benchmark datasets.
The new padded \gls{CNN} results are provided in the top portion of Table~\ref{tab:finn-key-results}, while key results from~\cite{finn} are shown in the lower portion.
Note that for comparison, scaled versions of the \glspl{MLP} consisting only of fully-connected layers described in \cite{binarynet} are also shown and denoted as \mlps{\sigma}.

We can see that larger networks scale well to larger \glspl{FPGA}, with our best designs achieving 14.8~\gls{TOPS} and 671~$\mu$s image classification latency.
Furthermore, even with the largest network tested, all model parameters fit within \gls{OCM} of the \LargeFpgaName{} and thus avoids potential bottlenecks on external memory access.
However, if we were to attempt a larger network (such as \cnns{2}) the design would no longer fit in \gls{OCM} without also reducing the frame rate.
This is discussed further in Section~\ref{subsec:bram}.

While the results described in Table~\ref{tab:finn-key-results} represent state-of-the-art in terms of image classification rates and energy efficiency, it is still work in progress.
Our best raw performance number (14.8~\gls{TOPS}) outperforms that of the smaller \gls{FPGA} device used in \OurScheme{}~\cite{finn} (11.6~\gls{TOPS}), which is no surprise.
However, the \glspl{MLP} shown in \cite{finn} do achieve performance figures closer to the theoretical peak of the device.
This is mostly due to the simplicity of \glspl{MLP} versus \glspl{CNN}.
Figure~\ref{fig:roofline} shows the estimated peak performance of the \LargeFpgaName{} with vertical lines indicating the arithmetic intensity of the 3 \gls{CNN} networks and coloured markers indicating actual performance of \OurScheme{}.
We can see that our implementations still fall below the \LargeFpgaName{}'s theoretical peak.
We expect that with planned improvements, including those in Section~\ref{subsec:bram}, significant performance gains can still be achieved.
However it should be noted, that the largest design \cnns{1} shown in Table~\ref{tab:finn-key-results} requires 1.2 \gls{GOP} per frame,
which is similar in computational requirements to the popular AlexNet~\cite{krizhevsky2012imagenet} which requires 1.45 \gls{GOP} per frame.
In comparison the GPUs, the NVidia Titan X can achieve 3.2 kFPS at 227 W for AlexNet inference, compared to 12 kFPS at less than 41 W on the KU115 \gls{FPGA}.%
\footnote{\detokenize{https://www.nvidia.com/content/tegra/embedded-systems/pdf/jetson_tx1_whitepaper.pdf}}
It should be noted that these figures are in terms of 32-bit floating point operations, as opposed to the binarized ones discussed in this work.
However, high accuracy has been achieved by fully binarized~\cite{hubara2016quantized} and partially binarized~\cite{dorefa} versions of AlexNet and we expect to be able to achieve high performance on such networks.

%

\subsubsection{BRAM Efficiency}
\label{subsec:bram}
\begin{figure}
	\centering
    \includegraphics{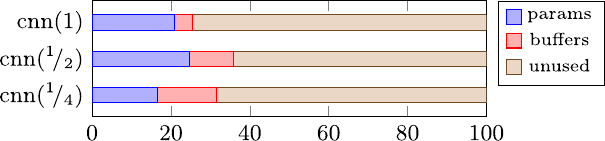}
	\caption{Utilization of allocated BRAM storage space.}
	\label{fig:BRAMUtil}
    \vspace{\figpad}
\end{figure}

Since FINN currently focuses on BNNs that fit entirely onto the on-chip memory of a single FPGA, making the most out of the available on-chip memory is essential.
Figure \ref{fig:BRAMUtil} illustrates how much of the allocated BRAM space (as reported by Vivado) is actually utilized by the accelerator.
The two largest contributors to BRAM usage in FINN are the network parameters (BNN weights and thresholds), and stream buffers (such as FIFOs and input-output buffers), which are shown with different colors in the bar chart.
As can be expected, the majority of the utilized storage is for weights, although the streaming buffers occupy roughly equal storage for \cnnf{1}{4} since there are not as many parameters.

A bigger concern is that on average only \approxtilde{22\%} of the storage space in the allocated BRAMs is actually used.
For scaling to even larger networks, this under--utilization could constitute a problem as synthesis will fail trying to allocate more BRAMs than is available in the FPGA.
Further analysis into this issue revealed that this is a consequence of how convolutions are currently handled in FINN.
Recall that the total folding factor is $F^{\mathrm{tot}}=F^s \cdot F^n \cdot F^m$ for a convolution layer.
The $F^m$ folding factor here arises due to implementing matrix--matrix products as a sequence of matrix--vector products
Unlike $F^s$ and $F^n$, $F^m$ is currently not controllable, since only one matrix--vector product is computed at a time in each MVTU.
When high FPS is desired, the initiation interval must be minimized, which can only be achieved by small values $F^n$ and $F^s$ since $F^m$ is constant.
This requires creating many PEs and SIMD lanes operating in parallel, each of which have their own weight and threshold memories operating independently.
However, this causes the weight matrix to be split and distributed into many small pieces, thus causing the observed storage under--utilization.

\begin{figure}
	\centering
	\includegraphics[width=0.6\linewidth]{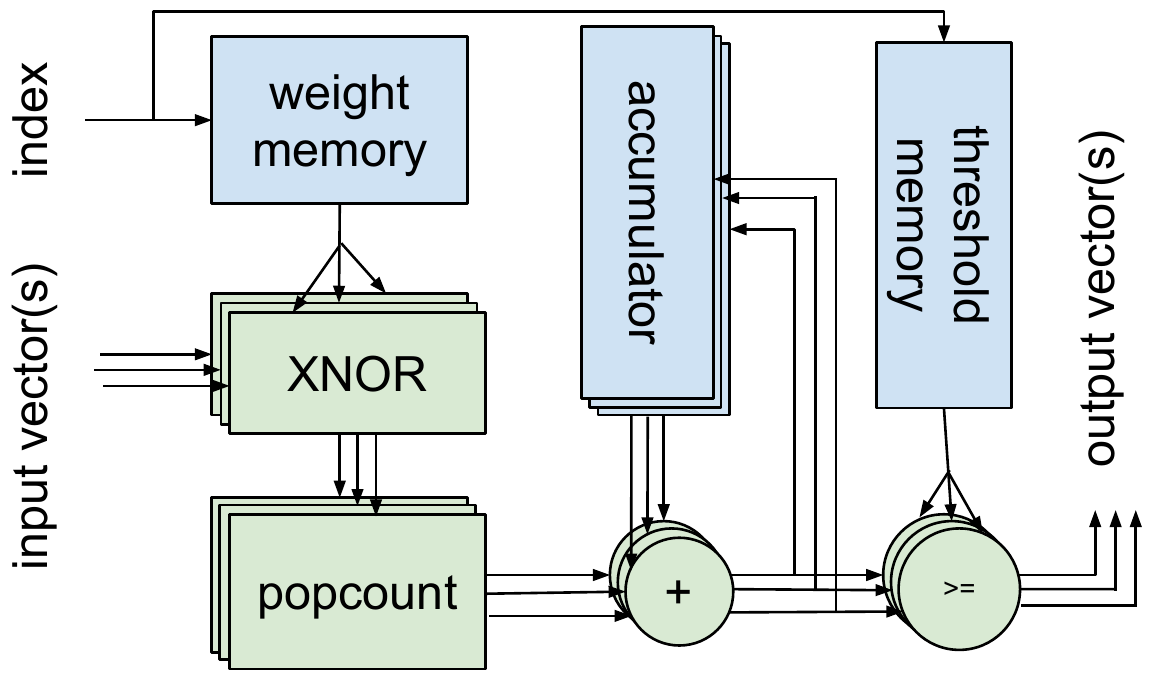}
	\caption{Datapath for matrix--multiple vector product.}
	\label{fig:MMVTU}
    \vspace{\figpad}
\end{figure}

One way of addressing this problem would be enabling control over the $F^m$ parameter by enhancing the \gls{MVU} to enable multiplying the same matrix by multiple vectors in parallel.
In this manner, fewer PEs and SIMD lanes could be instantiated, each working on a larger portion of the weight matrix and utilizing BRAM storage better.
Figure \ref{fig:MMVTU} shows how the \gls{MVU} datapath could be enhanced to support multiple vectors, broadcasting the same data from the weight memory to multiple XNOR-popcount-accumulate datapaths.
Note that only the datapath is duplicated; the weight and threshold memories have a single copy.
We leave further investigation of the matrix--multiple vectors for future work.

\section{Conclusion}
\label{sec:conclusion}

In this work, we explored the scaling of \glspl{BNN} on large \glspl{FPGA} using the \OurScheme{} framework.
We highlight an issue with padding in convolutional layers in \glspl{BNN} described in \cite{binarynet} which would cause them to require a 2-bit datapath.
We show that a small modification to padding (padding with -1 values) improves accuracy over no-padding and is comparable to 0-padding, while still allowing networks to maintain a binary datapath.
We found that high performance for large networks can be attained, with our highest demonstrated performance achieving 12 kFPS at less than 41 W of board power and 14.8 \gls{TOPS} of raw computational performance.
When scaling to large networks, we also show that the efficiency of BRAM usage in \OurScheme{} is low, and propose an architectural modification which would allow for better BRAM utilization.
Alternatively, if a higher number of smaller BRAMs were available on \glspl{FPGA} devices, this would allow \OurScheme{} to better exploit the available resources.

For future work, we will further enhance the \OurScheme{} framework to support partial binarization, and different kinds of convolutional layers, such as inception layers~\cite{szegedy2015going} and fire-modules~\cite{iandola2016squeezenet}.
The architectural improvements, described in Section~\ref{subsec:bram} will be implemented to further improve the BRAM usage efficiency of architectures produced by \OurScheme{}.
Further networks which have been trained on larger datasets, i.e., ImageNet, will also be implemented.
Finally, better power measurements will be attained rather than using ``worst-case'' power dissipation values.



\bibliographystyle{abbrv}
\bibliography{references}

\end{document}